\documentclass{article}

\usepackage[nonatbib,final]{neurips_2020_ml4ad}

\usepackage[utf8]{inputenc} % allow utf-8 input
\usepackage[T1]{fontenc}    % use 8-bit T1 fonts
\usepackage{hyperref}       % hyperlinks
\usepackage{url}            % simple URL typesetting
\usepackage{booktabs}       % professional-quality tables
\usepackage{amsfonts}       % blackboard math symbols
\usepackage{nicefrac}       % compact symbols for 1/2, etc.
\usepackage{microtype}      % microtypography
\usepackage{graphicx}
\usepackage{wrapfig}
\usepackage{color}
\usepackage{amsmath}

\newcommand{\pbf}[1]{\vspace{-1.5mm}\paragraph{\textbf{#1 }}}

\title{3D-LaneNet+: Anchor Free Lane Detection using a Semi-Local Representation}

\author{%
  Netalee Efrat \\
  \And
  Max Bluvstein \\
  \And
  Shaul Oron \\
  \AND
  Dan Levi \\
  \And
  Noa Garnett \\
  \And
  Bat El Shlomo \\
  \AND
  \And \\
  General Motors\\
  Advanced Technological Center Israel \\
  \scriptsize{\texttt{$\{$netalee.efratsela, max.bluvstein, shaul.oron, dan.levi, noa.garnett, batel.shlomo$\}$@gm.com}}\\
  \And \\
}

\begin{document}

\maketitle

\begin{abstract}
  % 3D-LaneNet+ is a camera-based DNN method for anchor free 3D lane detection. We follow recently proposed 3D-LaneNet and alter it in two important ways extending it to handle previously unsupported lane topologies such as splits, merges and short lanes, as well as to better handle complex road surface geometries. Our first change is introducing a semi-local, Bird Eye View (BEV), tile representation that breaks down lanes into simple lane segments learning a parametric model for each segment. The second change is adding a learnt deep feature embedding that is used to cluster the segments together into full lanes. This combination allows 3D-LaneNet+ to be anchor free and avoid using lane anchors and lane model fitting as in the original 3D-LaneNet. We demonstrate the efficacy of 3D-LaneNet+ using both synthetic and real world data. Results show significant improvement relative to the original 3D-LaneNet that can be attributed to better generalization to complex lane topologies, curvatures and surface geometries. 

3D-LaneNet+ is a camera-based DNN method for anchor free 3D lane detection which is able to detect 3d lanes of any arbitrary topology such as splits, merges, as well as short and perpendicular lanes. We follow recently proposed 3D-LaneNet, and extend it to enable the detection of these previously unsupported lane topologies. Our output representation is an anchor free, semi-local tile representation that breaks down lanes into simple lane segments whose parameters can be learnt. In addition we learn, per lane instance, feature embedding that reasons for the global connectivity of locally detected segments to form full 3d lanes. This combination allows 3D-LaneNet+ to avoid using lane anchors, non-maximum suppression, and lane model fitting as in the original 3D-LaneNet. We demonstrate the efficacy of 3D-LaneNet+ using both synthetic and real world data. Results show significant improvement relative to the original 3D-LaneNet that can be attributed to better generalization to complex lane topologies, curvatures and surface geometries. 
\end{abstract}

\section{Introduction}
\label{sec:intro}

Camera based 3D lane detection is a cardinal component in many autonomous driving related tasks such as trajectory planning, vehicle localization and map generation to name a few. 

Recently, Garnett et al. \cite{garnett20183dlanenet} proposed 3D-LaneNet, a camera-based 3D lane detection method which proposes two novel concepts for lanes detection. The first is a CNN architecture with integrated Inverse Perspective Mapping (IPM) to project feature maps to Bird Eye View (BEV), and the second is an anchor based representation which allows casting the lane detection problem to a single stage object detection problem. Following a similar concept as in SSD \cite{liu2016ssd} or RetinaNet \cite{retina_net}, each BEV column serves as an anchor which regresses the entire lane as a polyline. This imposes strong constrains on the detected lane geometry and topology, limiting this methods ability to detect lanes that are not roughly parallel to the ego vehicle direction of travel. Lanes starting further ahead, and other important non-trivial topologies such as junctions, can not be represented hence are not detected by this method.  

In this work, we follow recently proposed anchor free detectors such as FCOS \cite{fcos2019tian} and CenterNet \cite{centernet_zhou2019objects}, and suggest an anchor free extension to 3D-LaneNet we term  \textit{3D-LaneNet+}. Unlike 3D-LaneNet that uses the column based anchors to encapsulate prior information about the lanes structure (long and continuous), anchor-free detectors do not introduce such priors. Their basic paradigm is dividing the input to non-overlapping cells where each cell learns to detect the object that occupies that cell and estimate the object's attributes (e.g. center, dimensions, orientation). Lanes, however, are not compact objects with easily defined centers. Therefore, instead of predicting the entire lane as a whole, we detect small lane segments that lie within the cell and their attributes (position, orientation, height). In addition, we learn for each cell a global embedding that allows clustering the small lane segments together into full 3D lanes. Our suggested solution alleviates the assumption that all lanes should be represented as polylines. This enables detecting lanes of any arbitrary topology including splits, merges, short lanes and lanes perpendicular to the vehicle's direction of travel. Supporting these additional lane topologies improves the detection recall of 3D-LaneNet+, leading to a significant performance improvement relative to the original 3D-LaneNet.

\begin{figure}%[!htb]
\vspace{-2mm}
\centering
\includegraphics[width=\textwidth]{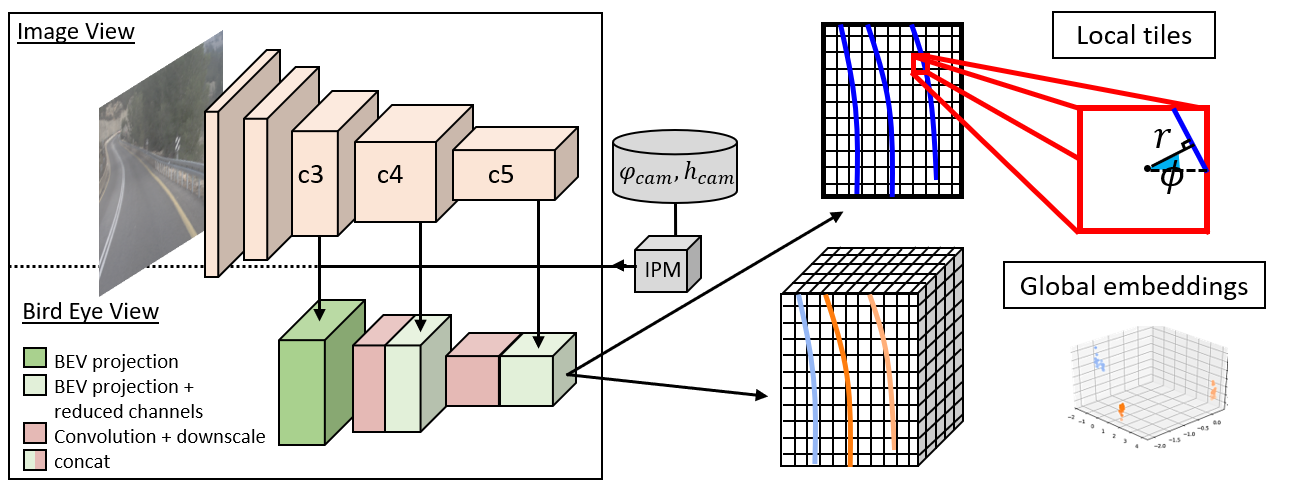}
\caption{Method overview. Our network is comprised of two processing pipelines: image view (top) and BEV (bottom). The final decimated BEV feature map is fed to the lane prediction head which outputs local lane segments and global embedding for clustering the segments to entire lanes curves.}
\label{fig:method}
%\vspace{-5mm}
\end{figure}

Our anchor free representation can be thought of as compact semi-local representation that is able to capture local topology-invariant lane structures and road surface geometries. Lane detection is done in Bird's Eye View using a grid of non-overlapping coarse tiles, as illustrated in Fig. \ref{fig:method}. We assume lane segments passing through the tiles are simple and can be represented by a low dimensional parametric model.  Specifically, each tile holds a line segment parameterized by an offset from the tile center, an orientation and a height offset from the BEV plane. This semi-local tile representation lies on the continuum between global representation (entire lane) to a local one (pixel level). Each tile output is more informative than a single pixel in a segmentation based solution as it is able to reason on the local lane structure but it is not as constrained as the global anchor based solution which has to capture together the complexity of the entire lane topology, curvature and surface geometry.

% Our semi-local representation breaks down lane curves into multiple lane segments but does not explicitly capture any relation between them. Adjacent tiles will have overlapping receptive fields, and thus correlated results, but the fact that several tiles represent the same lane entity is not captured. In order to generate full lane curves we learn an embedding for each tile which is globally consistent across the lane. This enables clustering small lane segments into full curves.

We run experiments, using both synthetic and real world datasets, that show 3D-LaneNet+ improves the mean average precision (MAP) over 3D-LaneNet by large margins. We demonstrate qualitatively and quantitatively the efficacy of our representation and its generalization to new lane curvatures and surface geometries.

\section{Related work}\label{sec:related_work}

\pbf{2D lane detection} Most existing lane detection methods focus on lane detection in the image plane and are mostly limited to detecting lanes parallel to the vehicle direction of travel. The literature is vast and includes methods performing 2D lane detection by using self attention \cite{Hou_2019_ICCV}, employing GANs \cite{ELGAN}, using new convolution layers \cite{pan2018SCNN}, exploit vanishing points to guide the training \cite{lee2017vpgnet} or use differentiable least-squares fitting \cite{wvangansbeke_2019}. Most related to ours is the method of \cite{DeepLearningHW_AndrewNg} that uses a grid based representation in the image plane, with a line parametrizations and density based spatial clustering for highway lane detection. Our approach performs 3D lane detection in BEV perspective, together with a learning based clustering as well as a different parametrization than \cite{DeepLearningHW_AndrewNg}. Another work related to ours is \cite{TowardsEnd2End} that uses learned embedding to perform lane clustering. While \cite{TowardsEnd2End} perform segmentation at the image pixel level, we cluster the lane segments in BEV on the semi-local tile scale, which is far less computationally expensive.

\pbf{3D lane detection} Detecting lanes in 3D is a challenging task drawing increasing attention in recent years. Some methods use LiDAR or a Camera and LiDAR combination for this task. For example, Bai et al.  \cite{bai2018deep} use a CNN over LiDAR points to estimate road surface height and then re-projects the camera to BEV accordingly. The network doesn't detect lane instances end-to-end, but rather outputs a dense detection probability map that needs to be further processed and clustered. 
More related to our work are camera-based methods. DeepLanes \cite{gurghian2016deeplanes} uses a BEV representation but works with top-viewing cameras that only detect lanes in the immediate surrounding of the vehicle without providing height information. 

Recently, Gen-LaneNet \cite{guo2020gen} suggested using the same column based anchor representation as in 3D-LaneNet \cite{garnett20183dlanenet}, but change the coordinate frame in which the 3d lane points are predicted. Although this new geometry guided lane anchor of Gen-LaneNet is more generalizable to unobserved scenes, it is still limited to long lanes that are roughly parallel to ego vehicle direction of travel. Our proposed anchor free representation is complementary to this new coordinate frame and is therefore expected to be useful in the context of Gen-LaneNet as well (although not tested in our work).  

%The closest related works to ours are 3D-LaneNet \cite{garnett20183dlanenet} and Gen-LaneNet \cite{guo2020gen}. Both of them use the column based anchor representation, while the difference between these methods resides in the coordinate frame in which the 3d lanes points are predicted. Although the new geometry guided lane anchor of Gen-LaneNet is more generalizable to unobserved scenes, it is still limited to long, roughly parallel to ego vehicle direction, lanes. In that sense, our anchor free representation is complementary to the coordinate frame hence would boost performance of both approaches. For practical reasons we chose 3d-LaneNet as our baselines. \netalee{Shaul, please revise if necessary}

\pbf{Anchor free detectors} Anchor-based methods detect objects in a one-shot global manner by representing the entire object using a set of parameters and regressing these parameters relatively to predefined anchors \cite{liu2016ssd, retina_net}. In the context of lane detection, this problem formulation constrains the lanes to be represented in a global manner by polylines, which limits the variety of lane topologies that can be predicted. Recent anchor free detection works alleviate the need for global anchors and enable the direct regression of the object parameters, or object's sub-parts parameters like in our case. These methods include CornerNet \cite{cornerNet_2018_ECCV} which detects objects as paired keypoints that correspond to the bounding box corners locations, CenterNet \cite{centernet_zhou2019objects} and FCOS \cite{fcos2019tian} which predicts the objects center and dimensions. AFDet, \cite{ge2020afdet} a LiDAR based detector, applies a similar method and predicts the center, orientation and dimensions of objects relatively to a regular top view grid. RepPoints \cite{yang2019reppoints} and Dense RepPoints \cite{yang2019densereppoints} detect objects by predicting sets of representative objects' points, taking a step towards non-rigid object representation which is more fitted to lanes. However, they predict the sets of points relative to the object center, whereas lanes do not have a well defined center. In 3D-LaneNet+ we represent the lane using a set of segments instead of points, thus incorporating information on the lane line structure, and instead of predicting the segments relative to a global lane center, we predict them relative to the grid cell centers. To reason about the segments connectivity we additionally learn instance embedding like in \cite{TowardsEnd2End}.

\begin{wrapfigure}{r}{0.5\textwidth}
\vspace{-3mm}
\includegraphics[width=0.48\textwidth]{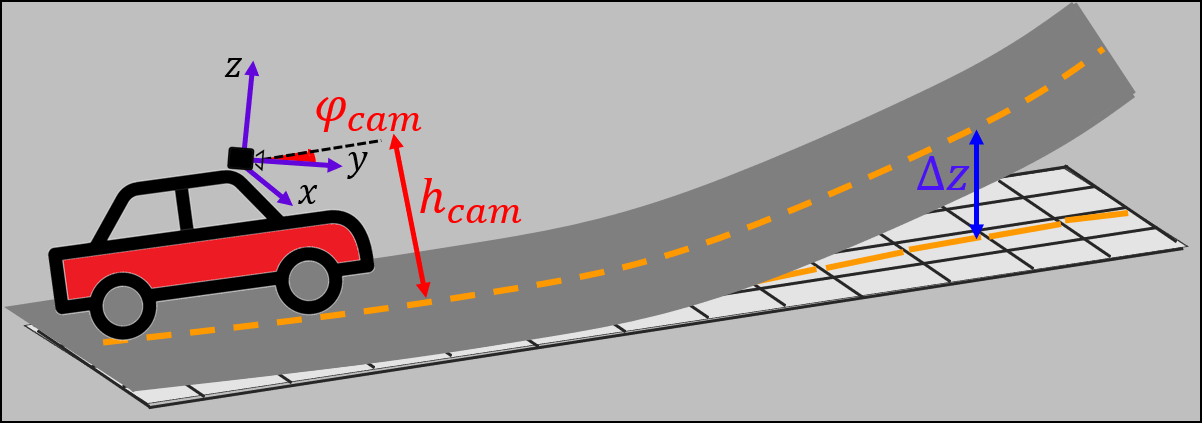}
\caption{The road projection plane is defined according to the camera mounting pitch angle $\varphi_{cam}$ and height $h_{cam}$, hence our representation is invariant to the camera extrinsics. We represent the GT lanes in full 3D relatively to that plane.}
\label{fig:projection_plane}
\vspace{-8mm}
\end{wrapfigure}
\section{Method}\label{sec:method}
We now describe our proposed 3D lane detection framework 3D-LaneNet+. A schematic overview appears in Fig.  \ref{fig:method}. We begin by presenting our semi-local tile representation and lane segment parameterization (Sec. \ref{sec:method_representation}) followed by how lane segments are clustered together using learnt embedding (Sec. \ref{sec:method_embedding}).

\subsection{Learning 3D lane segments with Semi-local tile representation}\label{sec:method_representation}

Lane curves have many different global topologies and lie on road surfaces with complex geometries. This makes reasoning for entire 3D lane curves a challenging task. Our key observation is that despite this global complexity, on a local level, lane segments can be represented by low dimensional parametric models. Taking advantage of this observation we propose a semi-local representation that allows our network to learn local lane segments thus generalizes well to unseen lane topologies, curvatures and surface geometries. 

%\begin{wrapfigure}{r}{0.5\textwidth}
%\includegraphics[width=0.48\textwidth]{projection_plane}
%\caption{The road projection plane is defined according to the camera mounting pitch angle $\varphi_{cam}$ and height $h_{cam}$, hence our representation is invariant to the camera extrinsics. We represent the GT lanes in full 3D relatively to that plane.}
%\label{fig:projection_plane}
%\end{wrapfigure}

Following Garnett et al. \cite{garnett20183dlanenet}, our network is given a single camera image that is fed to the dual pathway backbone which uses an encoder and an Inverse Perspective Mapping (IPM) module to project feature maps to BEV. The projection applies a homography, defined by camera pitch angle $\varphi_{cam}$ and height $h_{cam}$, that maps the image plane to the road plane (see Fig.  \ref{fig:projection_plane}). The final decimated BEV feature map is spatially divided into a grid $G_{W\times H}$ comprised of $W\times H$ non-overlapping tiles. The projection ensures each pixel in the BEV feature map corresponds to a predefined position on the road, independent of camera intrinsics and pose. 

We assume that through each tile $g_{ij}\in G_{W\times H}$ can pass a single line segment which can be approximated by a straight line. Specifically, the network regresses, per each tile $g_{ij}$, three parameters: lateral offset distance relative to tile center $\widetilde{r}_{ij}$, line angle $\widetilde{\phi}_{ij}$, (see Local tiles in Fig.  \ref{fig:method}) and height offset $\widetilde{\Delta z}_{ij}$ (see Fig.  \ref{fig:projection_plane}). In addition to these line parameters, the network also predicts a binary classification score $\widetilde{c}_{ij}$ indicating the probability that a lane intersects a particular tile. 
%GT regression targets for the offsets and angles are calculated by approximating the lane segments intersecting the tiles to straight lines using the GT lane points after they were projected to the road plane (Fig. \ref{fig:projection_plane}). Detailed derivation of the losses used can be found in the appendix (Sec. \ref{sec:tiles_loss}). 

Position and $z$ offsets are trained using an $L1$ loss: 
\begin{equation}
\mathcal{L}_{ij}^{Offsets} = \|\widetilde{r}_{ij} - r_{ij}\|_1 + \|\widetilde{\Delta z}_{ij} - \Delta z_{ij}\|_1
\end{equation}

Predicting the line angle $\widetilde{\phi}_{ij}$ is done using the hybrid classification-regression framework of \cite{Mahendran2018AMC} in which we classify the angle $\phi$ (omitting tile indexing for brevity) to be in one of $N_{\alpha}$ bins, centered at $\alpha = \{ \frac{2\pi}{N_{\alpha}}\cdot i \}_{i=1}^{N_{\alpha}}$. In addition, we regress a vector $\Delta^{\alpha}$, corresponding to the residual offset relative to each bin center. Our angle bin estimation is optimized using a soft multi-label objective, and the GT probabilities are calculated as $p^{\alpha} = [{1 - |\frac{2\pi}{N_{\alpha}}\cdot i - \phi| / \frac{2\pi}{N_{\alpha}}}]_+ $. 
% The GT offsets $\Delta^{\alpha}$ are the difference between the GT angle and the bin centers, and their training is supervised on the GT angle bin and adjacent bins to ensure that the delta offset can account for erroneous bin class prediction. 

The angle loss is the sum of the classification and offset regression losses:
\begin{multline}
 \mathcal{L}^{angle}_{ij} = \sum_{\alpha=1}^{N_{\alpha}}
     {[p^{\alpha}_{ij}\cdot \log{\widetilde{p}^{\alpha}_{ij}} + (1- p^{\alpha}_{ij}) \cdot \log{(1 - \widetilde{p}^{\alpha}_{ij})}  + \delta^{\alpha}_{ij} \cdot \|\widetilde{\Delta}^{\alpha}_{ij} - \Delta^{\alpha}_{ij} \|_1]}
\end{multline}
where $\delta^{\alpha}_{ij}$ is the indicator function masking the relevant bins for the offset learning.

The lane tile probability $\widetilde{c}_{ij}$ is trained using a binary cross entropy loss:
\begin{equation}
     \mathcal{L}^{score}_{ij} = c_{ij} \cdot\log{\widetilde{c}_{ij}} + (1-c_{ij}) \cdot\log{(1 - \widetilde{c}_{ij})}
\end{equation}

 Finally, the overall tile loss is the sum over all the tiles in the BEV grid:
\begin{equation}
     \mathcal{L}^{tiles} =  \sum_{i,j\in W\times H}( 
     \mathcal{L}^{score}_{ij} +c_{ij} \cdot \mathcal{L}^{angle}_{ij} +  c_{ij}\cdot \mathcal{L}^{offsets}_{ij})
     \label{Eq:loss_tiles}
\end{equation}

At inference we convert the offsets ($\widetilde{r}_{ij}, \widetilde{\Delta z}_{ij}$) and angles ($\widetilde{\phi}_{ij}$) back to points by converting them from Polar to Cartesian coordinates, and transform the points from the BEV plane to the camera coordinate frame by subtracting $h_{cam}$ and rotating by $-\varphi_{cam}$ (Fig. \ref{fig:projection_plane}).

\subsection{Global embedding for lane curve clustering}\label{sec:method_embedding}
Once lane tiles are obtained we are still left with the task of generating full 3D lane curves from these small segments. To this end we propose using a clustering approach based on learnt feature embedding. This technique removes the need for using lane anchors and polyline fitting as done in the original 3D-LaneNet, which in turn allow 3D-LaneNet+ to support previously unsupported lanes such as lane perpendicular to the ego vehicle direction of travel and short lanes that start at a long distance from the vehicle. 

Specifically, we learn an embedding vector $f_{ij}$ for each tile such that vectors representing tiles belonging to the same lane would reside close in embedded space while vectors representing tiles of different lanes would reside far apart. For this we adopted the approach of \cite{TowardsEnd2End, Semantic_Instance_Segmentation}, and use a discriminative push-pull loss. Unlike previous work, we use the discriminative loss on the decimated tiles grid, which is much more efficient  than operating at the pixel level. 

The discriminative push-pull loss is a combination of two losses:
\begin{equation}
 \mathcal{L}^{embedding} = \mathcal{L}^{pull} + \mathcal{L}^{push}
 \end{equation}
 A pull loss aimed at pulling the embeddings of the same lane tiles closer together:
\begin{equation}
     \mathcal{L}^{pull} = \frac{1}{C}\sum_{c=1}^C\frac{1}{N_c}\sum_{ij\in W\times H}{
     [\delta^c_{ij}\cdot\|\mu_c - f_{ij}\| - \Delta_{pull}]_+^2
     }
     \label{Eq:pull}
\end{equation}
 and a push loss aimed at pushing the embedding of tiles belonging to different lanes farther apart: 
\begin{equation}
     \mathcal{L}^{push} = \frac{1}{C(C-1)}\sum_{c_{A}=1}^C\sum_{c_{B}=1,c_{B}\neq c_{A}}^C{ [\Delta_{push} - \|\mu_{c_{A}} - \mu_{c_{B}}\|]^2_+ }
     \label{Eq:push}
\end{equation}

where $C$ is the number of lanes (can vary), $N_c$ is the number of tiles belonging to lane $c$, $\delta_{ij}^c$ indicates if tile $i,j$ belongs to lane $c$, $\mu_c=\frac{1}{N_c}\sum_{ij\in W\times H}{\delta^c_{ij}\cdot f_{ij} }$ is the average of $f_{ij}$ belonging to lane $c$, $\Delta_{pull}$ constraints the maximal intra-cluster distance and $\Delta_{push}$ is the inter-cluster minimal required distance. 

Given the learnt feature embedding we can use a simple clustering algorithm to extract the tiles that belong to individual lanes. We adopted the clustering methodology from Neven et al. \cite{TowardsEnd2End} which uses mean-shift to find the clusters centers and set a threshold around each center to get the cluster members. We set the threshold to $\frac{\Delta_{push}}{2}$.

%The input to our network is a single camera image. We adopted the dual pathway backbone proposed by Garnett et al. \cite{garnett20183dlanenet} which uses an encoder and an Inverse Perspective Mapping (IPM) module to project feature maps to BEV. The projection applies a homography, defined by camera pitch angle $\varphi_{cam}$ and height $h_{cam}$, that maps the image plane to the road plane (see Fig.  \ref{fig:projection_plane}). The final decimated BEV feature map is spatially divided into a grid $G_{W\times H}$ comprised of $W\times H$ non-overlapping tiles. Similar to \cite{garnett20183dlanenet}, the projection ensures each pixel in the BEV feature map corresponds to a predefined position on the road, independent of camera intrinsics and pose.

\section{Experimental Setup}\label{Sec:Experiments}
We study the performance of 3D-LaneNet+ using two 3D-lane datasets and compare it to the original 3D-LaneNet of Garnett et al. \cite{garnett20183dlanenet}. We show the advantage of our anchor free approach in detecting short lanes and lanes starting far from the host vehicle demonstrating the method ability to detect different lane topologies and generalize to complex surface geometries.  

\pbf{Datasets } Evaluation is done using two 3D lane datasets. The first is \textit{synthetic-3D-lanes} \cite{garnett20183dlanenet} containing synthetic images of complex road geometries with 3D ground truth lane annotations. The second, is a dataset we collected and annotated referred to as \textit{Real-3D-lanes}. This dataset (annotated as in \cite{garnett20183dlanenet}), contains $327K$ images from 19 distinct recordings (different geographical locations which spans an area of $250km$, at different times) taken at 20 fps. The data is split such that the train set, $298K$ images, is comprised mostly of highway scenarios while the test set is comprised of a rural scenario with complex curvatures and road surface geometries, taken at a geographic location not in the train set. 
To reduce temporal correlation we sampled every 30'th frames giving us a test set of 1000 images. Examples from the train and test sets can be seen in Fig. \ref{fig:real_data}.

\pbf{Evaluation} We adopt the same evaluation protocol used in the original 3D-LaneNet \cite{garnett20183dlanenet} work that proposes to separate the detection accuracy from the geometric estimation accuracy. Detection accuracy is computed via the commonly used Mean Average Precision (MAP) metric. Similarly to \cite{DAGMapper_2019_ICCV} we adopted the IOU measure to associate between detected curves and GT curves which is general and can account for short lanes or non-parallel lanes that do not go through predefined y-values. The geometric accuracy is assessed by measuring the lateral error of each lane point with respect to its associated GT curve. We divided the entire dataset to lane points in the near range (0-30m) and far range (30-80m) and calculate the mean absolute lateral error for every range.

\pbf{Implementation details } We use the dual-pathway architecture  \cite{garnett20183dlanenet} with a ResNet34 \cite{ResNet} backbone. Our BEV projection covers 20.4m x 80m divided in the last decimated feature map to our tile grid $G_{W\times H}$ with $W=16, H=26$ such that each tile represents $1.28m \times 3m$  of road surface. We found that predicting the camera angle $\varphi_{cam}$ and height $h_{cam}$ gave negligible boost in performance compared to using the fixed mounting parameters on \textit{Real-3D-lanes}, however, on \textit{synthetic-3D-lanes} we followed \cite{garnett20183dlanenet} methodology and trained the network to output $\varphi_{cam}$ and $h_{cam}$ as well. Our average runtime is 85.3 msec using a single GPU (NVIDIA Quadro P5000). 

The network is trained with batch size 16 using ADAM optimizer, with initial lr of 1e-5 for 80K iterations which is then reduced to 1e-6 for another $50K$ iterations. We set $\Delta_{pull}$ and $\Delta_{push}$ (Eqs. \ref{Eq:pull}, \ref{Eq:push}) to 0.1 and 3 respectively, and used a coarse 0.3 threshold on the output segment scores $\widetilde{c}_{ij}$ prior to the clustering. %In the evaluation we set a coarse distance threshold for association of $1m$, and measured the AP as the average of $AP_{IOU\%}$ at IOU thresholds $0.1:0.1:0.9$.  

\section{Results}\label{sec:results}

\begin{figure}[ht]
%\vspace{-3mm}
\centering
\includegraphics[width=\textwidth]{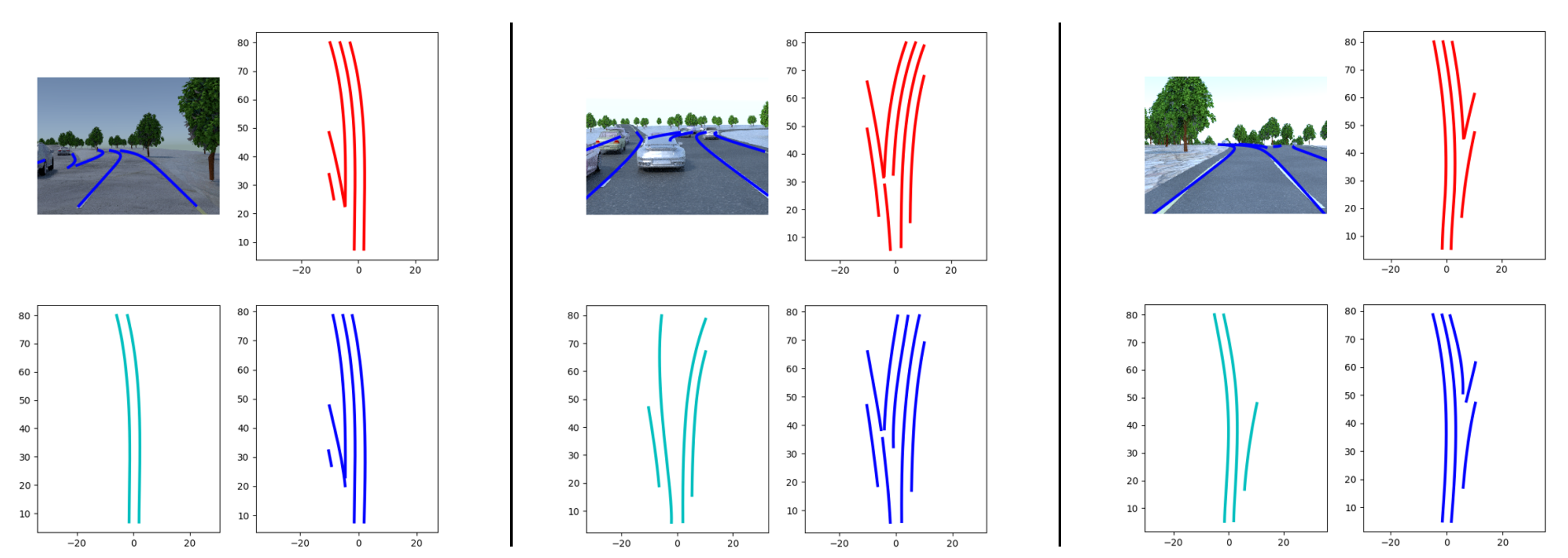}
\caption{Example results on \textit{synthetic-3D-lanes}.Detected lanes using 3D-LaneNet+ (Our)  \textcolor{blue}{(blue)}, ground truth \textcolor{red}{(red)} and 3D-LaneNet \cite{garnett20183dlanenet} \textcolor{cyan}{(cyan)}. It can be seen that due to the use of lane anchors 3D-LaneNet \textcolor{cyan}{(cyan)} misses many short lanes, splits and lanes starting far ahead from the ego vehicle. Out proposed anchor free approach allows 3D-LaneNet+ \textcolor{blue}{(blue)} to be less constrained and detect splits, short lanes and far lanes that were missed by 3D-LaneNet.}
\label{fig:blender15}
%\vspace{-4mm}
\end{figure}

\pbf{Synthetic-3D-lanes dataset }
We compared our 3D-LaneNet+ method to the original 3D-LaneNet. Results are presented in Table \ref{Table:blender}. It can be seen that our $MAP$ (mean of $AP_{IOU\%}$ at IOU thresholds $0.1:0.1:0.9$) and $AP_{50}$ are superior to those of 3D-LaneNet\footnote{Results reported here differ than those in \cite{garnett20183dlanenet} since their evaluation disregards short lanes that start beyond 20m from the ego vehicle.} while showing comparable lateral error (for $IOU=0.5$ and $recall=0.75$). We believe the main reason is our semi-local representation that allows our method to support many different lane topologies such as short lanes, splits and merges that emerge only at a certain distance from the ego vehicle. This is evident in Fig. \ref{fig:blender15} showing examples where splits and short lanes are not detected by 3D-LaneNet but detected by 3D-LaneNet+. 

\begin{table}[!htbp] 
%\vspace{-2mm}
\centering
\caption{comparison on \textit{Synthetic-3D-lanes} dataset}
\label{Table:blender}
\begin{tabular}{||l||cc||c|cc||}
\hline
Method & $MAP$ &  $AP_{50}$ & recall &\multicolumn{2}{c||}{Lateral error (cm)} \\
\cline{5-6}
 & & & & 0-30m &  30-80m\\
\hline
3D-LaneNet \cite{garnett20183dlanenet}   &  0.74 & 0.79  &0.75 &\textbf{9.3}  & \textbf{23.9} \\
3D-LaneNet+ (Ours)   &  \textbf{0.9} & \textbf{0.95}  &0.75 & 9.7  & 26.7 \\
\hline
\end{tabular}
%\vspace{-2mm}
\end{table}

\pbf{Real world data} We use the \textit{Real-3D-lanes} dataset to demonstrate our methods ability to handle real world data while generalizing well to different lane topologies, curvatures and surface geometries. Results comparing 3D-LaneNet+ to 3D-LaneNet, trained on the same train set, are summarized in Table \ref{Table:real_data}. 

\begin{figure}[ht]
%\vspace{-2mm}
\centering
\includegraphics[width=\textwidth]{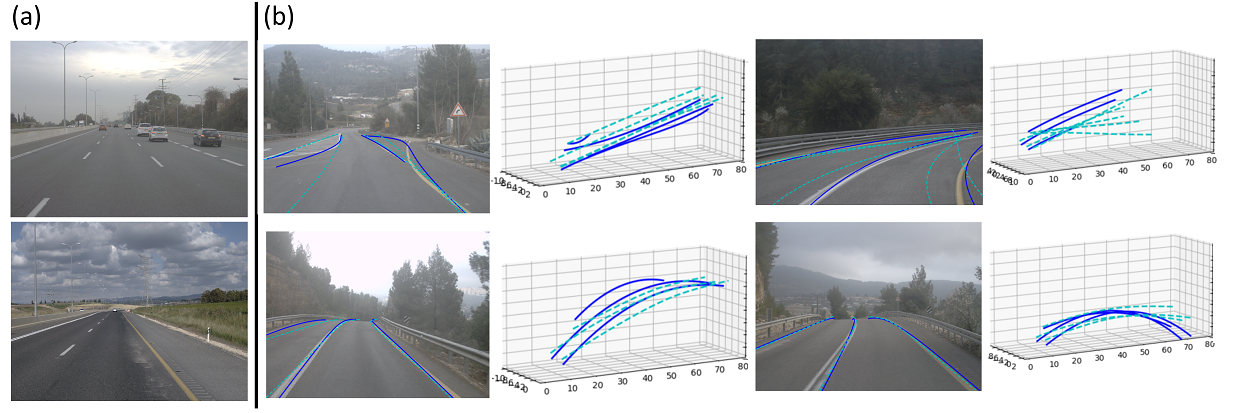}
\caption{Example results on \textit{Real-3D-lanes}. (a) Sample images from the training set, (b) examples from the test set. Lanes detected using 3D-LaneNet+ shown in \textcolor{blue}{blue} and lanes detected by 3D-LaneNet in \textcolor{cyan}{cyan}.It can be seen that 3D-LaneNet+ produces more accurate results both by better capturing the correct lane curvature as well as fitting the surface geometry more tightly. Removing the use of lane anchors also helps in reducing the number of false detections.}
\label{fig:real_data}
%\vspace{-2mm}
\end{figure}

It can be seen that 3D-LaneNet+ achieves better results improving by 9 points over 3D-LaneNet in overall $MAP$ as well as lowering the lateral error for the 3d lane points. This experiment is challenging compared to the \textit{synthetic-3D-lanes} experiment in which train and test sets have the same distribution. In the case of \textit{Real-3D-lanes}, train and test sets were recorded in different geographic location, and have different distributions, i.e., the test set exhibits more complex curvatures and surface geometries. To support this claim we examined the distribution of lane curvature and road surface curvature. On the test set, lane curves have, on average, a curvature score that is higher by an order of magnitude from the train set. The road surface curvature is two orders of magnitude higher in the test set than on the train set (see Fig. \ref{fig:real_data}b for example images). Our ability to generalize to this test set demonstrates the advantage of using the proposed semi-local tiles representation.

\begin{table}[!htbp] 
\vspace{-2mm}
\centering
\caption{comparison on \textit{Real-3D-lanes} dataset}
\label{Table:real_data}
\begin{tabular}{||l||ccc||c|cc||}
\hline
Method  & $MAP$ &  $AP_{50}$ & $AP_{90}$&  Recall & \multicolumn{2}{c||}{Lateral error (cm)}\\
\cline{6-7}
 & & & & & 0-30m &  30-80m\\
\hline

3D-LaneNet \cite{garnett20183dlanenet}   &  0.80 & 0.86 & 0.48   & 0.85 & 15.6  & 47.2 \\
3D-LaneNet+ (Ours) w/o global  &  0.84 & 0.94 & 0.43 & 0.85 & 14.5  & 45.5 \\

3D-LaneNet+ (Ours) &  \textbf{0.89} & \textbf{0.95} & \textbf{0.60}  & 0.85 &\textbf{14.1}  & \textbf{44.7} \\

\hline
3D-LaneNet+ (Ours) w synthetic  & \textbf{0.9}  & \textbf{0.95} & 0.59 & 0.85 & \textbf{12.9}  & \textbf{36.3} \\
\hline
\end{tabular}
% \vspace{-4mm}
\end{table}

To show the significance of our clustering approach using global feature embedding we compare it with a na\"ive clustering alternative (Table \ref{Table:real_data} '3D-LaneNet+ w/o global'). This alternative uses a simple greedy algorithm concatenating segments based on continuity and similarity heuristics. We find that when using na\"ive clustering we loose 5 points in overall $MAP$ and 17 in $AP_{90}$ suggesting that detected lanes with greedy clustering are much shorter. In addition, we see that with feature embedding we obtain lower lateral error. This may suggest that feature embedding learning also helps predicting more accurate segments. 

We also compare the model trained only on \textit{Real-3D-lanes} with a one trained on both \textit{Real-3D-lanes} and \textit{synthetic-3D-lanes} (Table \ref{Table:real_data} '3D-LaneNet+ w synthetic'). We find that additional 3D training data of complex curvatures and geometries helps in the generalization despite it being synthetic and without using any domain adaptation techniques.

\begin{figure}[h]
% \vspace{-2mm}
\centering
\includegraphics[width=\textwidth]{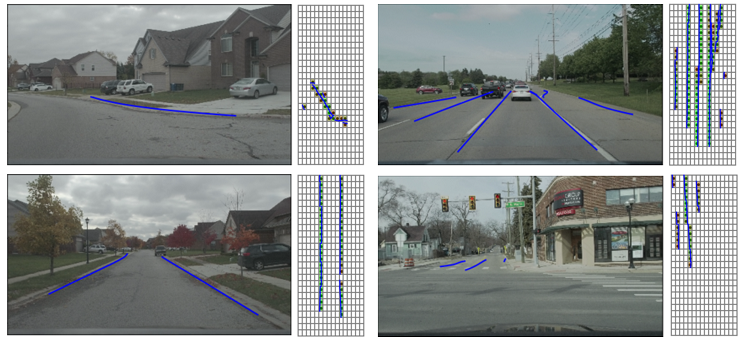}
\caption{generalization to a new camera and scenes: examples from our \textit{internal evaluation dataset}. Our method generalizes well both to the new camera that was not present in the training set, as well as to new scenes, without any need for adaptation. On the right of each example are the detected lane segments depicted on the tiles grid. Color represent the predicted segment confidence, higher is \textcolor{green}{green}, lower is \textcolor{red}{red}.}
\label{fig:generalization}
% \vspace{-2mm}
\end{figure}

\pbf{Generalization to new cameras} We now examine our method's generalization to a new unseen camera. To this end, we conduct a qualitative evaluation on an \textit{internal evaluation dataset} which was not used for training. We can see from Fig. \ref{fig:generalization} that our method succeeds in detecting lanes from unseen camera without any adaptation needed. The examples in Fig. \ref{fig:generalization} shows our representation ability to support different topologies as well as to generalize to new scenes. From upper left clockwise one can see perpendicular lane, split, short lanes starting only across the intersection and a suburbs scene with only road edges that was not present in the train set at all. Note that training our model on these examples, including urban scenes with junctions and perpendicular lanes, will obviously achieve better results while the former methods 3D-LaneNet \cite{garnett20183dlanenet} and Gen-LaneNet \cite{guo2020gen} won't benefit from such training data because of their constrained representation.

%More experiment on variants of our method are presented in Table \ref{Table:real_data}. The first experiment comes to test the importance of the global feature embedding component. Removing the feature embedding training enforces us to find an alternative clustering mechanism. To this end, we used a na\"ive clustering approach which greedily concatenates predicted lane segments to form entire lanes based on continuity and similarity heuristics. We can see that when using the greedy clustering we loose in $AP$, especially in $AP_{90}$ which indicates that the detected lanes with the greedy clustering are much shorter. In addition, we can see that the lateral error is lower with the feature embedding than without. This is interesting because it can suggest that the feature embedding not only provides better clustering, but it helps predicting more accurate segments. 
%Another interesting comparison to make is between our method trained only on \textit{3D-lanes} and our method trained on both \textit{3D-lanes} and \textit{synthetic-3D-lanes}. It is visible that providing the network additional 3D training data of complex curvatures and geometries helps significantly in the generalization ability of our method, even tough this data is synthetic and we didn't use any domain adaptation techniques.
\section{Conclusions}
In this work we introduced 3D-LaneNet+ a 3D lane detection framework that builds upon and improves the recently published 3D-LaneNet. Our main goal was extending 3D-LaneNet allowing it to detect lane topologies previously not supported, including short lanes, lanes perpendicular to the ego vehicle, splits, merges and more. 
In order to improve performance in these challenging cases we developed an anchor free method by introducing a semi-local representation that captures topology-invariant lane segments that are then clustered together using a learned global embedding into full lane curves. Removing the need to use lane anchors and lane model fitting (as in 3D-LaneNet) allows 3D-LaneNet+ to support and generalize to different lane topologies, curvatures and surface geometries. 
The efficacy of 3D-LaneNet+ was demonstrated on both synthetic and real world data producing a significant improvement in 3D lane detection compared to 3D-LaneNet. A qualitative inspection of the results shows that indeed the method is better equipped to successfully detect lanes of arbitrary topology including splits, merges, short lanes and more. By doing so we take another step towards meeting the requirements of 3D lane detection in autonomous driving in both highway and urban scenarios.

\clearpage
\bibliographystyle{ieee}
\bibliography{bibliography.bib}
\clearpage

\end{document}